\documentclass[aps,preprint,showpacs,superscriptaddress]{revtex4-1}  % for double-spaced preprint
\usepackage{graphicx}
\usepackage{subcaption}
\usepackage{bm}   
\usepackage{color}
\usepackage{amssymb}
\usepackage{amsmath}
\usepackage{float}
\begin{document}
	
	\title{Prediction of chaotic attractors in quasiperiodically forced logistic map using deep~learning}
	\author{J.Meiyazhagan}
	\affiliation{Department of Nonlinear Dynamics, Bharathidasan University, Tiruchirappalli - 620 024, Tamilnadu, India}
	\author{M. Senthilvelan}
	\email[Correspondence to: ]{velan@cnld.bdu.ac.in}
	\affiliation{Department of Nonlinear Dynamics, Bharathidasan University, Tiruchirappalli - 620 024, Tamilnadu, India}
	\vspace{10pt}
	
	\begin{abstract}
	We forecast two different chaotic dynamics of the quasiperiodically forced logistic map using the well-known deep learning framework Long Short-Term Memory. We generate two data sets and use one in the training process and the other in the testing process. The predicted values are evaluated using the metric called Root Mean Square Error and visualized using the scatter plots. The robustness of the Long Short-Term Memory model is evaluated {using} the number of units in the layers of the model. {We also} make multi-step forecasting of the considered system. We show that the considered Long Short-Term Memory model performs well in predicting chaotic attractors upto three steps.
	\end{abstract}
	
	%
	% Uncomment for keywords
	%\vspace{2pc}
	%\noindent{\it Keywords}: XXXXXX, YYYYYYYY, ZZZZZZZZZ
	%
	% Uncomment for Submitted to journal title message
	%\submitto{\JPA}
	%
	% Uncomment if a separate title page is required
	%\maketitle
	% 
	% For two-column output uncomment the next line and choose [10pt] rather than [12pt] in the \documentclass declaration
	%\ioptwocol
	%
	\maketitle
\section{Introduction}
\par Recently, Machine Learning (ML) and Deep Learning (DL) models have been used in various fields of physics \cite{Carleo2019,sudhe1,sudhe3}. In the study of dynamics of nonlinear systems, ML and DL algorithms are extensively used for the prediction and discovery of the behaviour of the chaotic and complex systems. For example, they have been used to identify chimera states \cite{BARMPARIS2020,ganaie2020}, in the replication of chaotic attractors \cite{Pathak2017}, using symbolic time series for network classification \cite{panday2021}, separating chaotic signals \cite{Krishnagopal2020}, learning dynamical systems in noise \cite{santo1} and in the prediction of extreme events \cite{meiyazhagan2021,meiyazhagan2021-2,PYRAGAS2020,dibak1,dibak2,asch2021model}. Very recently, the authors of Ref.~\cite{lellep2020} have considered H\'enon map and used a ML algorithm, namely Artificial Neural Network (ANN), to study the extreme events in it. The authors have focussed on binary classification and classified the data points as extreme and non-extreme \cite{lellep2020}. 
\par
\begin{figure}[!ht]
	\centering
	\includegraphics[width=0.8\linewidth]{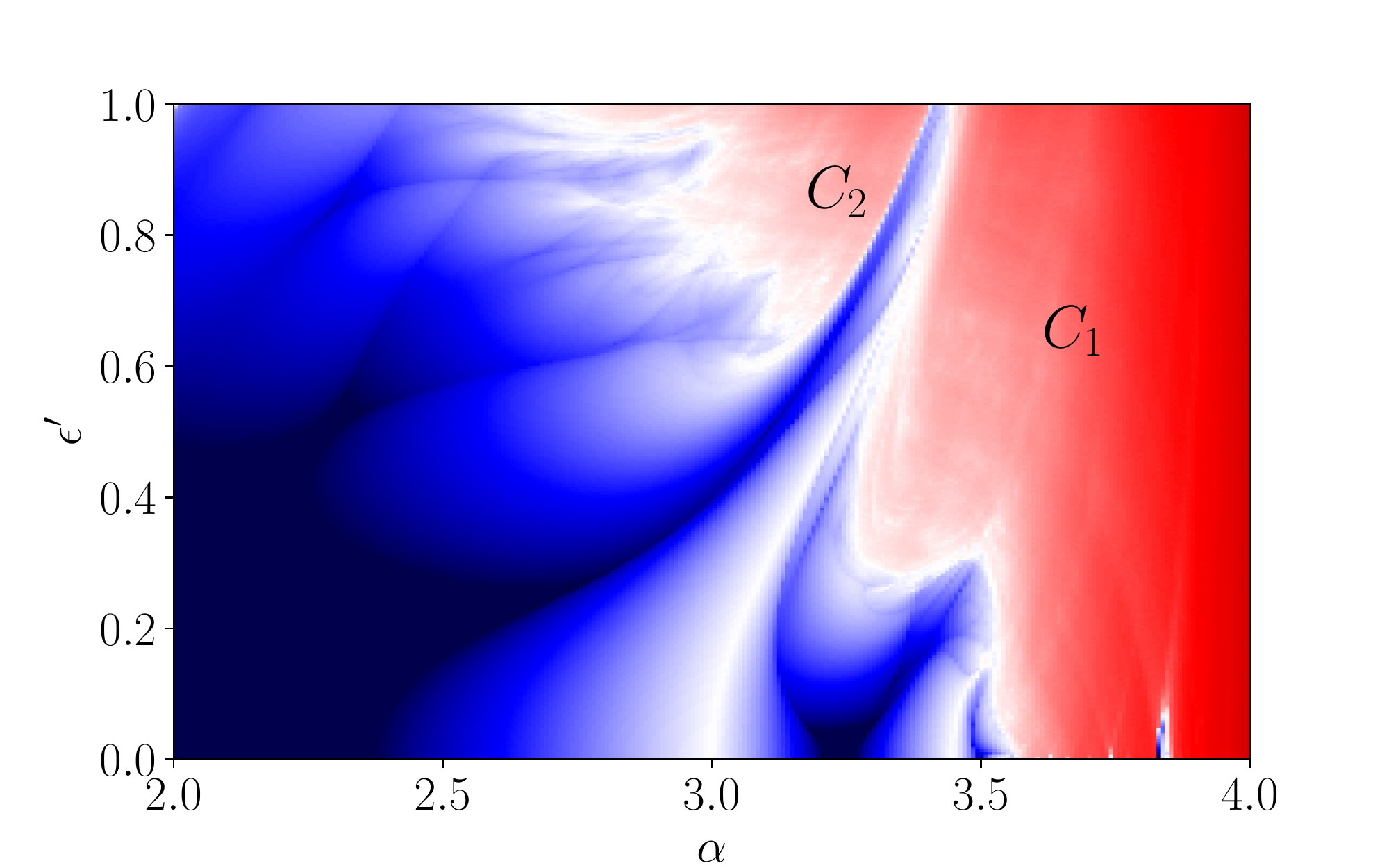}
	\caption{\label{fig:phase} Schematic phase diagram of the quasiperiodically forced logistic map. $C_1$ and $C_2$ are two different chaotic regimes.}
\end{figure}

 In our studies, we consider logistic map with quasiperiodic forcing and predict the time series of the system which is not continuous. The system exhibits chaos in two different regimes. We predict both the chaotic attractors of this system with the help of the DL framework, namely Long Short-Term Memory (LSTM). The logistic map with quasiperiodic forcing is described by the following equations, namely \cite{prasad1998,heagy1994}
\begin{subequations}\label{map}
	\begin{equation}
		x_{n+1} = \alpha [1+\epsilon\cos(2\pi\phi_n)]x_n(1-x_n),
	\end{equation}
	\begin{equation}
		\phi_{n+1}=\phi_n+\omega\; (\textrm{mod 1}),
	\end{equation}
\end{subequations}
where $\epsilon$ and $\omega=(\sqrt{5}-1)/2$ are the forcing amplitude  and irrational driving frequency respectively. The authors in Ref.~\cite{prasad1998} redefined the driving parameter as $\epsilon ' =\epsilon/(4/\alpha-1)$ to study the dynamics of the system in the regimes of $0\leq x \leq 1$, $0\leq \phi \leq 1$ and $0\leq \epsilon \leq 1$. The schematic phase diagram \cite{prasad1998} of the system is given in Fig.~\ref{fig:phase}. 
The system shows various dynamic behaviours, namely periodic, strange nonchaotic and chaotic attractors which can be characterized by the nonzero Lyapunov exponent $\Lambda$ \cite{prasad1998}, where
\begin{equation}
	\Lambda = \lim_{N\rightarrow\infty}\dfrac{1}{N}\sum_{i=1}^{N}\ln|\alpha[1+\epsilon\cos(2\pi\phi_i)](1-2x_i)|.
\end{equation} 
From Fig.~\ref{fig:phase}, we can {notice} the interesting behaviour of the considered system which has two chaotic regimes, namely $C_1$ and $C_2$. The $C_1$ regime is the continuation of the chaotic regime in the logistic map for $\epsilon=0$ at the end of the period-doubling cascade, at $\alpha=3.5699...$. {The chaos }in $C_2$ regime is due to low nonlinearity and large amplitude forcing \cite{prasad1998}. Our aim is to predict chaotic attractors in both the regime using LSTM model since it is capable of forecasting the data which is in the form of {a} sequence.
\par We organize our work as follows. In Sec. 2, we discuss the generation of training and testing data. In Sec. 3, we consider a DL framework called LSTM and train it using training set data and predict the test set data. The performance of the LSTM model is discussed in Sec. 4. We present the conclusion in Sec. 5.
\section{Data preparation}
\par
\begin{figure}[!ht]
	\centering
	\includegraphics[width=1.0\textwidth]{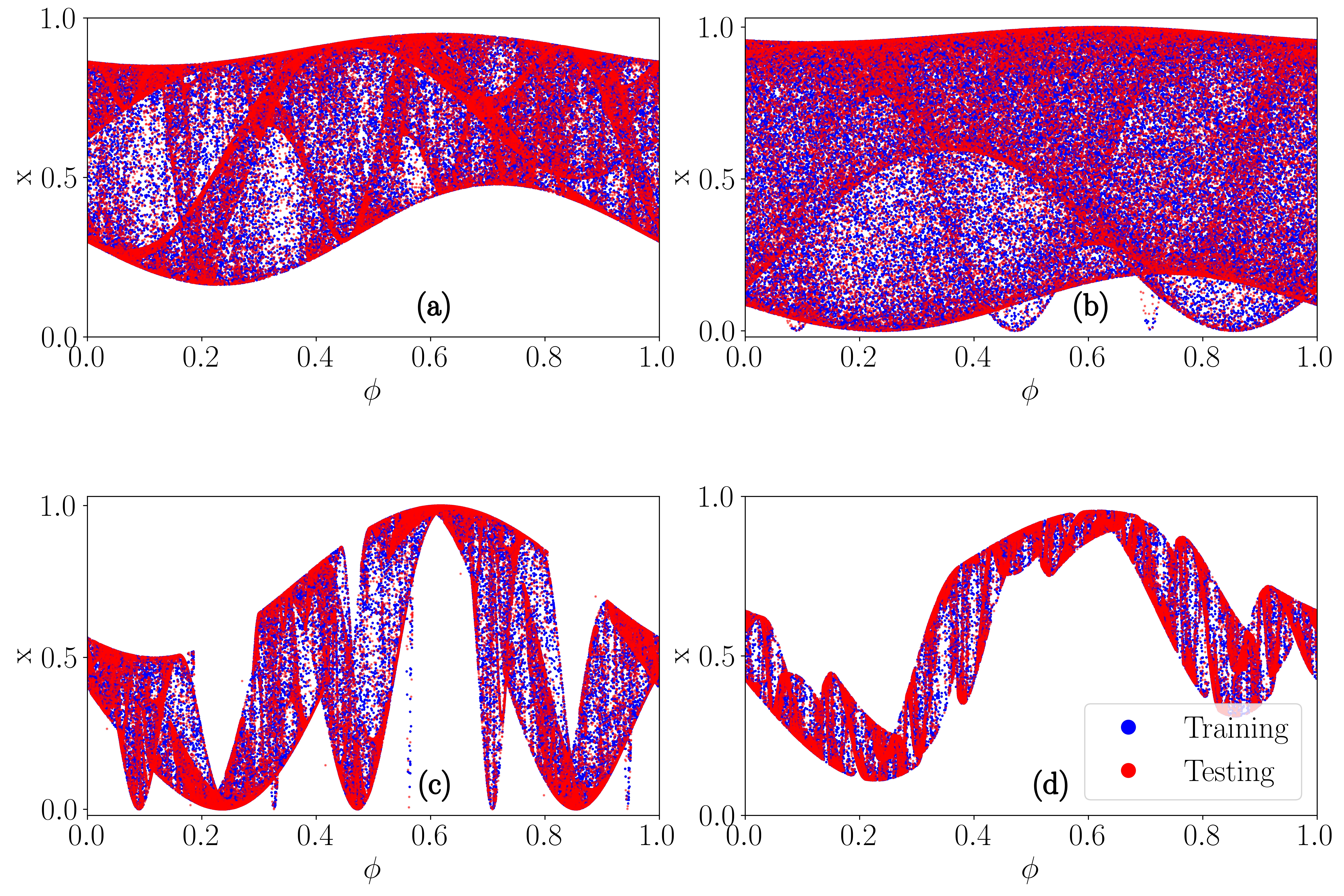}
	\caption{\label{fig:tt} Chaotic attractors in two different regimes $C_1$ and $C_2$. (a) $\alpha=3.6, \epsilon '=0.5$ and (b) $\alpha=3.9, \epsilon '=1.0$ correspond to $C_1$. (c) $\alpha=3.0, \epsilon '=1.0$  and (d) $\alpha=3.1, \epsilon '=0.8$ correspond to $C_2$. The points in blue (colour online) denoting the training set data and red (colour online) denoting the test set data.}
\end{figure}

 Generating data is the foremost task in prediction because prediction is done only by learning the relationship between the given data. We calculate the value of $x$ for $10^5$ iterations using equation \eqref{map} in both the regimes $C_1$ and $C_2$.  This discrete space data is then converted into supervised learning data by taking $x_n$ as input and $x_{n+1}$ as output. The chaotic attractors in the both regimes $C_1$ and $C_2$ are shown in Fig.~\ref{fig:tt}. The Figs.~\ref{fig:tt} (a) \& (b) corresponding to the regime $C_1$ and Figs.~\ref{fig:tt} (c) \& (d) correspond to the $C_2$ regime. 
The values of the parameters are taken as (a) $\alpha=3.6, \epsilon '=0.5$, (b) $\alpha=3.9, \epsilon '=1.0$, (c) $\alpha=3.0, \epsilon '=1.0$  and (d) $\alpha=3.1, \epsilon '=0.8$. We divide the data into two parts: (i) training set and (ii) test set. Training set data are used during the training process of the DL model and test set data are used for the evaluation of the ability of the DL model. In Fig.~\ref{fig:tt}, the blue dots are the data used for training purpose and the red coloured data are used for testing. We use $6\times 10^4$ data as training set data and $4\times 10^4$ data as test set data.

\par These two sets of data are rescaled using min-max normalization which is given by the formula \cite{al2006data},
\begin{equation}
	x_i^{rescaled} = a + \dfrac{(x_i-x_{min})(b-a)}{x_{max}-x_{min}}, \qquad i = 1,2,3,\hdots,n,
\end{equation}
where $x_{min}$ and $x_{max}$ are the minimum and maximum value of the data set respectively. We fix $a=-1$ and $b=+1$ {in order to scale} the data between $-1$ {and} $+1$. During the testing phase, this preprocessing scaling step is reversed after {obtaining the} output from the DL model in order to compare the results with the actual data.

\section{Deep Learning framework: Long Short-Term Memory}
\par When the data is in a sequential form one can make use of the Recurrent Neural Networks (RNN) \cite{rumelhart1986} which is a type of ANN. For the present study we consider a DL framework known as LSTM \cite{hochreiter1997} which is a special kind of RNNs. In recent years, LSTM framework has proven to be capable of forecasting time series of the chaotic systems even when there are extreme events in the time series \cite{meiyazhagan2021,dibak1,dibak2}. The main feature that differentiates LSTM from the other RNNs is that the latter has only one activation function for the neurons that is $\tanh$ but in the case of the former, a sigmoid function is used for recurrent activations and $\tanh$ is used for the activation of neurons. The sigmoid activation function is defined by \cite{goodfellow2016},
\begin{equation}
	\sigma (z) = \dfrac{1}{1+e^{-z}}.
\end{equation}
\par We construct the LSTM model in the following way. We consider two LSTM layers each having 16 units in it and followed by a layer which has one neuron for output. During the training, we give both {the} input and the corresponding output to the model, that is {we give $x_n$ as the input and $x_{n+1}$ as the output}. By doing this, the model will learn the nonlinear relations between the given data. After training, the learned model is used to forecast the data steps. 
\begin{figure}[!ht]
	\centering
	\includegraphics[width=1.0\linewidth]{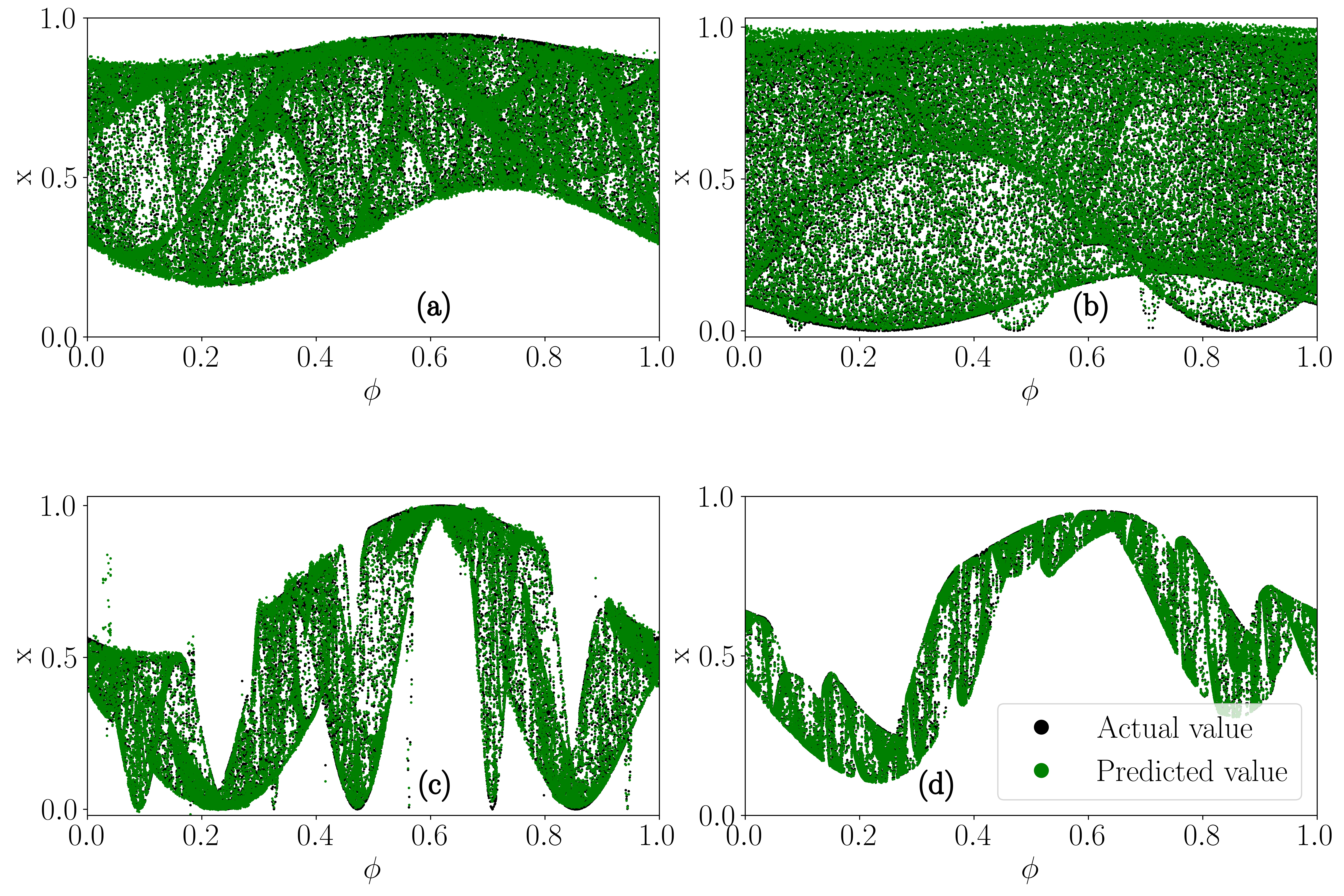}
	\caption{\label{fig:main_results} Plots of forecasted values over the actual values for four different sets of $\alpha$ and $\epsilon$ values as mentioned in Fig.~\ref{fig:tt}. The Figs.~(a) \& (b) correspond to $C_1$ regime and (c) \& (d) correspond to $C_2$ regime. Black dots denote the actual value and green dots denote the predicted value.}
\end{figure}
During the testing phase, we feed only the input data and ask the model for the corresponding output. The predicted values at the output given by the LSTM model are compared with actual values to determine the efficiency of the model in forecasting the chaotic attractors of the considered system.
\begin{figure}[!ht]
	\centering
	\includegraphics[width=0.8\linewidth]{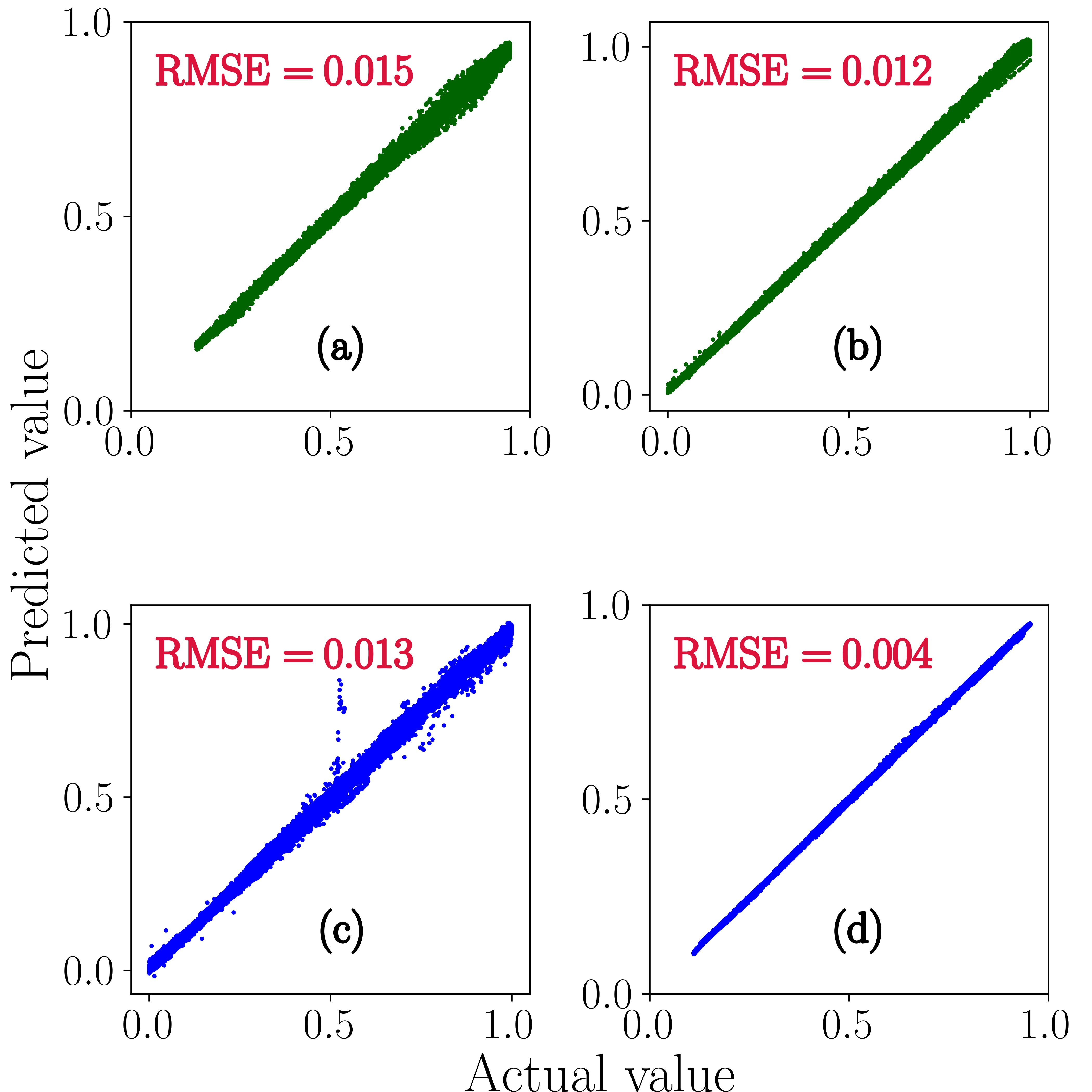}
	\caption{\label{fig:main_scatter} Scatter plots and RMSE values of the four different cases. The green colour plots correspond to the regime $C_1$ and blue colour corresponds to $C_2$.}
\end{figure}

\section{Results and Discussion}
\par To visualize the performance of the considered DL model, predicted data are plotted over the actual data in Fig.~\ref{fig:main_results}. Black dots (colour online) denote the actual value and green dots (colour online) denote the predicted value. The data in Figs.~\ref{fig:main_results} (a) \& (b) correspond to $C_1$ regime and Figs.~\ref{fig:main_results} (c) \& (d) correspond to $C_2$ regime. From the plots, we can see that relatively all predicted data coincide with the actual data. To have a clear understanding of the efficiency of the model we calculate the Root Mean Square Error (RMSE) value using the formula,
\begin{equation}
	RMSE = \sqrt{\sum_{i=1}^{N_{Test}}\dfrac{(\hat{Y}_i^{Test}-Y_i^{Test})^2}{N_{Test}}},
\end{equation}
where $\hat{Y}_i^{Test}$, $Y_i^{Test}$ and $N_{Test}$ {denote the} predicted values, actual values and total number of data in the test set {respectively}. We make use of the scatter plots which are plotted by taking actual values in the $x$-axis and predicted values in the $y$-axis (see Fig.~\ref{fig:main_scatter}). From Figs.~\ref{fig:main_scatter} (a) \& (b) we can see that the RMSE values for the regime $C_1$ are $0.015$ and $0.012$ respectively for the parameter values $\alpha=3.6, \epsilon '=0.5$ and $\alpha=3.9, \epsilon '=1.0$. The outcome of the scatter plots almost fit in straight line, {thereby} indicating that the difference between predicted and actual values are very low. From Figs.~\ref{fig:main_scatter} (c) \& (d) we can see that the the results of the second regime $C_2$ are calculated as $0.013$ and $0.004$ respectively for the parameter values $\alpha=3.0, \epsilon '=1.0$  and  $\alpha=3.1, \epsilon '=0.8$. The scatter plots for the test set data of regime $C_2$ {also} show very {little} scatter points, {thereby} indicating the best fit of predicted data with the actual data.

\begin{figure}[!ht]
	\centering
	\includegraphics[width=0.9\linewidth]{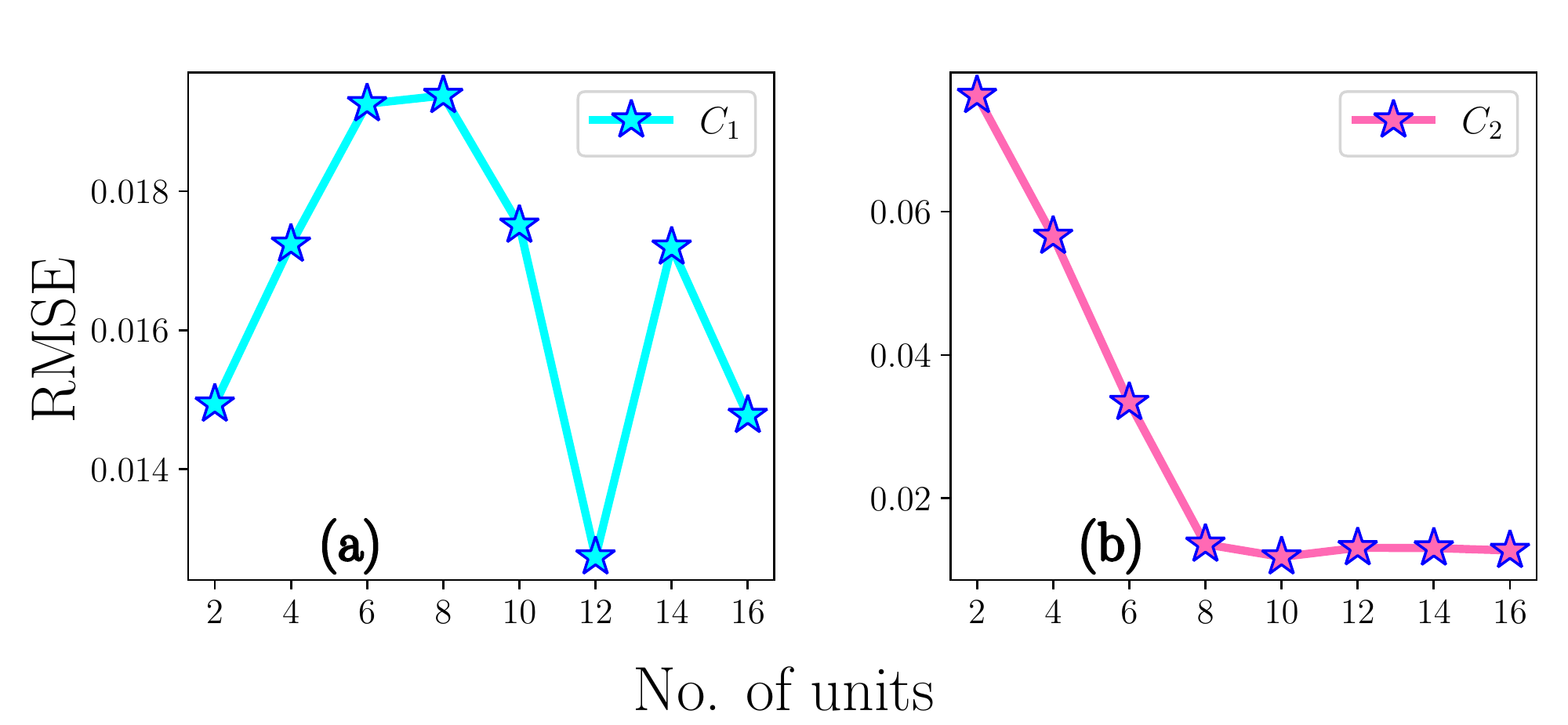}
	\caption{\label{fig:units} RMSE values for various number of units in LSTM layers. (a) and (b) corresponds to the regimes $C_1$,  $\alpha=3.6, \epsilon '=0.5$ and $C_2$, $\alpha=3.0, \epsilon '=1.0$ respectively.}
\end{figure}
\subsection{Effect of model architecture}
\par To study the effect of model architecture on the performance of the considered model we vary the number of units and analyse the performance based on the RMSE values. For this purpose, we change the units in both LSTM layers and train the model. Then each trained model is evaluated using the test set data. The outcome is shown in Fig.~\ref{fig:units}. For the $C_1$ regime, we evaluate the model with the data corresponding to $\alpha=3.6, \epsilon '=0.5$ and plot the results in Fig.~\ref{fig:units} (a). For the $C_2$ regime, we evaluate the model with the data corresponding to $\alpha=3.0, \epsilon '=1.0$ and plot the results in Fig.~\ref{fig:units} (b). The RMSE value changes while varying the number of units in the LSTM layers. 

\subsection{Multi-step forecasting}
\par
\begin{figure}[!ht]
	\centering
	\includegraphics[width=1\linewidth]{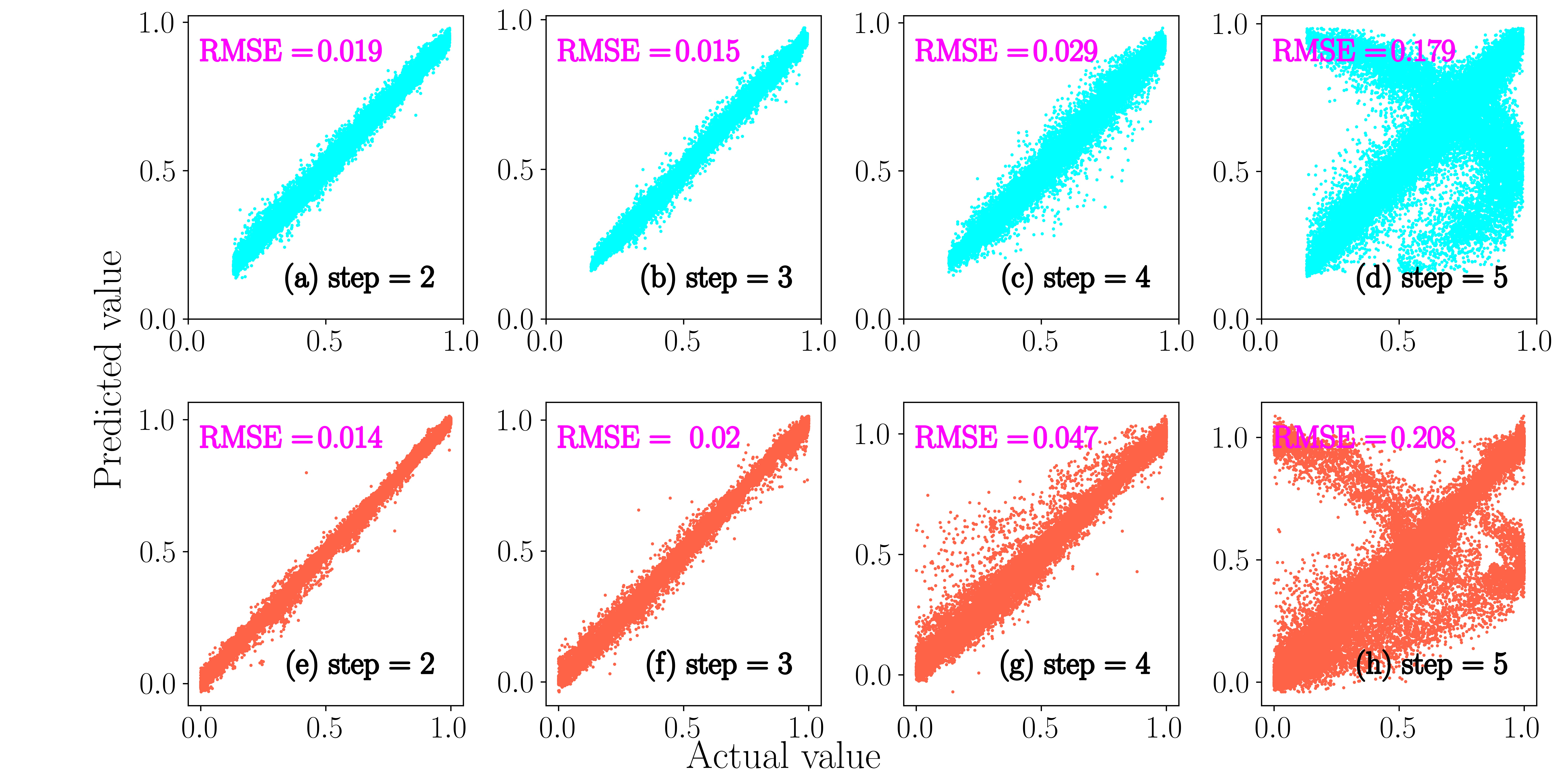}
	\caption{\label{fig:sctt_step} Scatter plots with RMSE values for the multi-step forecasting. (a)-(d) correspond to $C_1$ regime $\alpha=3.6, \epsilon '=0.5$ and (e)-(f) correspond to $C_2$ regime $\alpha=3.0, \epsilon '=1.0$.}
\end{figure}

 Now, we consider the task of multi-step forecasting. To do this, while preparing the supervised learning data, instead of having only one future step value, we take more than one value at the output. For this we consider the data in both the regimes $C_1$ ($\alpha=3.6, \epsilon '=0.5$) and $C_2$ ($\alpha=3.0, \epsilon '=1.0$). The results of multi-step forecasting are shown in Fig.~\ref{fig:sctt_step}. 
From this figure we can infer that in the forecasting of multi-steps two and three, the considered model outperformed our expectations in the prediction task, the plots have fewer scatter points and the RMSE values are in admissible range. But for steps four and five, the model failed to give accurate values in both the regimes. It can be seen from Figs.~\ref{fig:sctt_step} (c), (d), (g) and (h), the points are scattered much when making forecasting with fourth and fifth steps.
\section{Conclusion}
\par In this work, we have considered the logistic map with quasiperiodic forcing. The system exhibits chaos in two different regimes. We employed a DL framework LSTM, for the prediction of two different chaos. For this, we have generated $10^5$ data totally and used $6\times 10^4$ data for training and the remaining $ 4\times 10^4$ data for the purpose of testing. We forecast the chaos corresponding to the two regimes $C_1$ and $C_2$. The outcome of the experiments are evaluated using the performance metric RMSE value and they are analyzed through the scatter plots which {have been} plotted between the predicted value and actual value. Further, we have checked the effect of the number of units of the LSTM layers on the performance of the model. In this connection, we have done multi-step forecasting in order to predict more than one future value of the considered map. From the obtained results, we conclude that the developed LSTM framework can be used for forecasting the chaotic dynamics of the discrete system, namely quasiperiodically forced logistic map described by Eq.~\eqref{map}.

\section*{Acknowledgements}
JM thanks RUSA 2.0 project for providing a fellowship to carry out this work. MS acknowledges RUSA 2.0 project for providing financial support in procuring a high-performance GPU server which highly assisted this work.

\bibliography{mybibfile.bib}
\end{document}